\documentclass[letterpaper, 10 pt, conference]{ieeeconf}  

\IEEEoverridecommandlockouts                              

\usepackage{graphicx}
\usepackage{kotex}
\usepackage{tabularx}
\usepackage{adjustbox}
\usepackage{dsfont}
\usepackage{url}
\usepackage{color}
\usepackage{cite}
\usepackage{bbm}
\usepackage{array}
\usepackage[table]{xcolor}
\usepackage{booktabs}
\usepackage{bbding}
\usepackage{pifont}
\usepackage{amssymb}
\usepackage{tabularray}
\usepackage{colortbl}
\usepackage{caption}
\usepackage{multirow}
\usepackage{nicematrix}
\usepackage{float}

\title{\LARGE \bf
How to Relieve Distribution Shifts in Semantic Segmentation for Off-road Environments}

\author{Ji-Hoon Hwang, Daeyoung Kim, Hyung-Suk Yoon, Dong-Wook Kim and Seung-Woo Seo
\thanks{All authors are with Department of Electrical and Communication Engineering, Seoul National University, Korea
        {\tt\small hoons21@snu.ac.kr}}%
}

\begin{document}
\captionsetup[table]{position=bottom}
\captionsetup[figure]{skip=3pt}

\maketitle
\thispagestyle{empty}
\pagestyle{empty}

\begin{abstract}
Semantic segmentation is crucial for autonomous navigation in off-road environments, enabling precise classification of surroundings to identify traversable regions. However, distinctive factors inherent to off-road conditions, such as source-target domain discrepancies and sensor corruption from rough terrain, can result in distribution shifts that alter the data differently from the trained conditions. This often leads to inaccurate semantic label predictions and subsequent failures in navigation tasks.
To address this, we propose ST-Seg, a novel framework that expands the source distribution through style expansion (SE) and texture regularization (TR). Unlike prior methods that implicitly apply generalization within a fixed source distribution, ST-Seg offers an intuitive approach for distribution shift.
Specifically, SE broadens domain coverage by generating diverse realistic styles, augmenting the limited style information of the source domain. TR stabilizes local texture representation affected by style-augmented learning through a deep texture manifold. Experiments across various distribution-shifted target domains demonstrate the effectiveness of ST-Seg, with substantial improvements over existing methods. These results highlight the robustness of ST-Seg, enhancing the real-world applicability of semantic segmentation for off-road navigation.

\vspace{-0.1cm}
\end{abstract}

\section{INTRODUCTION}
\begin{figure}[t]
      \includegraphics[width=1\linewidth]{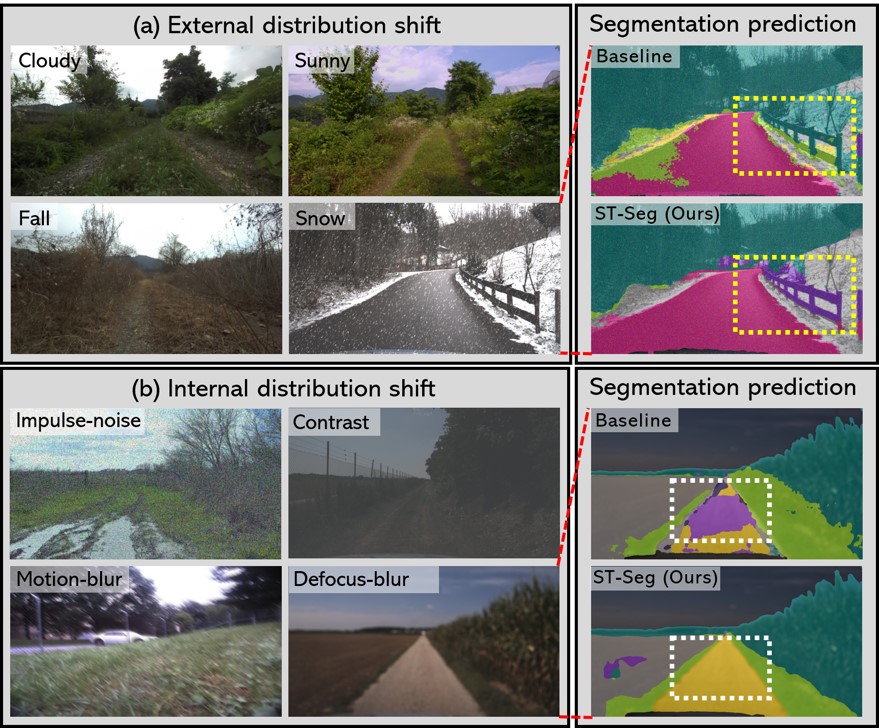}
      \captionsetup{font=small}
      \caption{\textbf{Performance of ST-Seg on various distribution-shifted target domains.} Examples of (a) external and (b) internal distribution shifts during off-road navigation are shown. The yellow box marks cases where the baseline fails to detect obstacles, while the white box shows misclassification of traversable terrain as obstacles.}
      \vspace{-0.8cm}
      \label{intro fig}
   \end{figure}

Research on off-road mobile robot navigation has gained significant attention in recent years due to its diverse applications, including transportation \cite{arte}, disaster detection \cite{disaster}, and agricultural robotics \cite{agriculture}. 
A key component of these applications is semantic segmentation \cite{hao2020brief}, a representative task in robot perception area that assigns a semantic label to each pixel, offering critical navigation information not provided by geometric sensor data. 
However, due to the unstructured and highly variable nature of off-road environments, previous semantic segmentation method often fail to maintain performance, leading to incorrect predictions in real-world scenarios \cite{ganav, offseg, sansaw} (Fig. \ref{intro fig}).
Such inaccurate perception results can lead to critical damage to the robot, particularly in off-road environments where rough terrain and unpredictable obstacles are prevalent \cite{arte}.

In this paper, we observe that this issue arises from a phenomenon known as \textit{distribution shift}, which occurs in deep learning when the environmental conditions encountered during deployment differ from those of the training dataset \cite{distributionshift}.
We conceptualize this problem from two perspectives: \textit{external distribution shift} and \textit{internal distribution shift}. The \textit{external distribution shift} stems from discrepancies between the pre-acquired source domain and the target domain encountered during robot operation. This is prominent in off-road environments, where unstructured objects like natural elements lack fixed shapes and vary across locations and times (Fig. \ref{intro fig}-(a)). The \textit{internal distribution shift} results from corrupted sensor data as the robot navigates unstructured off-road environments. Unlike urban settings, where robots traverse smooth roads, off-road environments force robots to move over rough terrain, significantly impacting sensors with noise and blur (Fig. \ref{intro fig}-(b)). Thus, internal shift can occur even in the source domain that has already been learned.

In the context of supervised learning, the \textit{distribution shift} arises due to the limited \textit{style} information in the source domain used for training. The \textit{style} \cite{adain} refers to the feature-level statistics that capture the global appearance of an image, a key characteristic of a domain.
Since the model cannot access any information about the target domains during training, \cite{sansaw} removed style information to learn domain-agnostic features. While this approach reduces domain-specific biases, it also weakens feature representation power. To address this, \cite{domainrandom1} introduced random style adjustment and applied style-augmented learning by training with style-augmented features. However, random style adjustment generates unrealistic styles, leading to inconsistencies with real-world styles, and since this random approach cannot be said to effectively expand distribution coverage, it cannot be considered an explicit solution to distribution shift.
We also observe that altering style information disrupts \textit{texture} \cite{stylecontent2}, which represents the feature embedding values capturing an image's local patterns and is partially influenced by style \cite{stylecontentorthogonal}.

Due to these issues, previous semantic segmentation methods still fail to maintain their performance in distribution-shifted target domains.

In this paper, we propose ST-Seg based on our observations to relieve distribution shifts by incorporating Style Expansion (SE) and Texture Regularization (TR) methods.
To address inconsistencies caused by non-realistic, randomly generated styles, SE improves adaptability to distribution-shifted target domains by performing style-augmented learning with realistic and diverse generated styles, leading to broader domain coverage.
Specifically, the style generator utilizes the extensive style information from the ImageNet \cite{imagenet} to learn a distribution model of realistic styles.
The style sampler subsequently employs stratified sampling \cite{stratified} to extract diverse, unbiased styles, enabling the generation of extensive style-augmented features.
Additionally, to mitigate the instability of local texture information that inevitably arises when adjusting global style information, TR regularizes texture information using a deep texture manifold obtained from natural texture data \cite{gtos}, thereby maintaining a consistent representation of natural textures and improving the robustness of style-augmented learning.

Contributions of this paper are summarized as follows:

\begin{itemize}

\item We propose an explicit solution to mitigate distribution shifts by expanding the source distribution itself, in contrast to previous methods that implicitly apply generalization techniques within a fixed source distribution.

\item We propose style expansion (SE), which broadens domain coverage through style augmentation learning, and texture regularization (TR), which ensures consistent texture feature representation. Together, these methods enhance the robustness of semantic segmentation models under distribution shifts.

\item We demonstrate that ST-Seg consistently maintains robust performance under both internal and external distribution shifts, validated through diverse off-road datasets, including challenging real-world scenarios collected from our UGV platform.

\end{itemize}

\section{Related Works}

\begin{figure*}[thpb]
      \centering
      \includegraphics[width=0.9\linewidth]{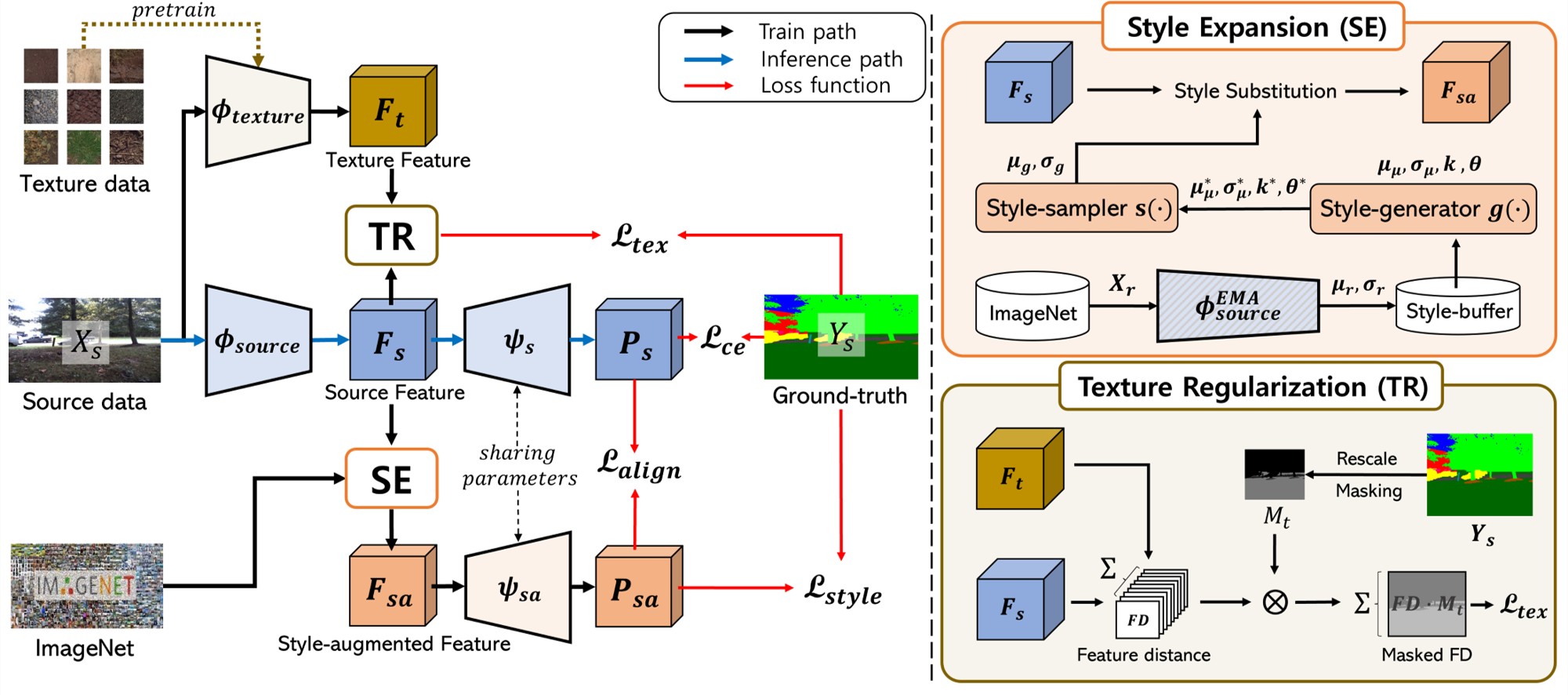}
      \captionsetup{font=small}
      \caption{ \textbf{The overall framework of ST-Seg.} The left diagram illustrates how SE and TR are applied throughout the entire architecture. The right diagram is a block representation to facilitate understanding of the SE and TR methods individually. A detailed explanation of the figure can be found in the subsection \ref{Overview}.}
      \vspace{-0.4cm}
      \label{mainfig}
   \end{figure*}

\subsection{Semantic Segmentation in Off-road Environments}
It is important to understand the navigation characteristics and traversability of the terrain for off-road navigation \cite{arte}.
The advancement of deep learning networks and the emergence of datasets from various off-road environments have led to extensive research in semantic segmentation for off-road environments \cite{rugd, rellis, tas, deepscene, ycor, goose}.
From early CNNs \cite{bisenetv2, deeplabv3+, mobilenet} to transformer-based networks \cite{segformer, ganav}, various architectures have been employed to off-road environments, with focusing on efficient designs to enable real-time operation.
However, it is crucial to develop specialized training strategies designed for distribution shift of unstructured nature in off-road environments, rather than depending solely on existing validated segmentation networks.

Recent studies have applied additional modules to extract robust features, allowing the model to implicitly address the distribution shift problem during the training phase \cite{uncertainty, ganav}.
However, this implicit approach still fails to maintain performance on distribution-shifted target domains, as it provides guidance to the model using only source data that is not exposed to distribution shifts.
We propose ST-Seg, which changes the approach by explicitly utilizing domain-specific information, termed style, to expand the known domain coverage of the model and enable robust predictions in the distribution-shifted target domain.

\vspace{-0.05cm}
\subsection{Style Information in Semantic Segmentation}
The channel-wise mean and standard deviation of the feature map, known as style \cite{adain}, is a fundamental element representing a specific domain. 
Numerous studies have been utilized style information to enhance the generalization performance of semantic segmentation.
\cite{batch} adjusts the style of all batch normalization (BN) layers in the network to match the style of target domain, preventing performance degradation in the target domain.
While this works fine when target domain is accessible, however in off-road navigation settings, access to the target domain is not feasible.

To address a more practical problem where the target domain cannot be accessed, \cite{sansaw} have adopted instance normalization (IN) based methods, which separates style information during the training process to learn domain-agnostic features.
While using IN to transform all samples into a standard normal distribution during training phase can improve the generalization performance of the model, it weakens the feature representation power because the style information is removed. 
To overcome the limitations of IN based methods, instead of removing the styles during the training phase,  \cite{domainrandom1} diversified the distribution of style through random style adjustments; however, this random style may introduce disparities with real-world style. 
Moreover, style-augmented learning approaches that alter style may introduce variations in texture information \cite{stylecontentorthogonal}, resulting in unstable learning of feature representations.
To address these issue, we propose ST-Seg which can expands the limited style distribution of the source domain in a realistic and unbiased manner and stabilize the learning of feature representation.
\vspace{-0.07cm}
\section{Methodology}

\subsection{Problem Formulation}
\noindent \textbf{Task Settings}
The source domain dataset \(D_{s}=\left\{(X_{s},Y_{s})\right\}\) refers to the training data for which we have class labeled ground truth, while the target domain \(D_t=\{(X_t)\}\) refers to the dataset that has distribution shift and is inaccessible during the training process. 
\(X_{s}, X_t\in \mathbb{R}^{H\times W\times 3}\) is an image, \(Y_{s}\in \mathbb{R}^{H\times W\times C}\) is pixel-level semantic label, \(C\) is the number of semantic classes, \(H\) and \(W\) are the height and width of the image.
Our goal is to train the source encoder \(\phi_{source}\) and decoder \(\psi_{s}\) to better predict pixel-level class probability maps \(P_s\in \mathbb{R}^{H\times W \times C}\) across multiple distribution-shifted target domains \(D_t\) while learning solely on the source domain \(D_s\).

\noindent \textbf{Style and Texture}
The channel-wise mean and standard deviation of the feature map, known as style information, is a global information representing a specific domain \cite{adain}. 
Following this works, let \(F_{s}^{l}\in\mathbb{R}^{d\times H_{l}\times W_{l}}\) be the \(l\)-th layer feature of \( X_{s}\) from the source encoder network \(\phi_{source}\), where \(l\) denotes the number of layer in encoder and \(d\) denotes the dimension of channels. The style statistics \(\mu_{s}^l\in\mathbb{R}^{d}\), \(\sigma_{s}^l\in\mathbb{R}^{d}\) of the feature \(F_{s}^{l}\) can be calculated as follows:
\begin{equation}
\label{statistics1}
\textstyle
\mu_{s}^{l} = \frac{1}{H_{s}W_{s}}\sum_{h=1}^{H_{l}}\sum_{w=1}^{W_{l}}F_{s}^{l},
\end{equation}
\begin{equation}
\label{statistics2}
\textstyle
\sigma_{s}^{l} = \sqrt{\frac{1}{H_{l}W_{l}}\sum_{h=1}^{H_{l}}\sum_{w=1}^{W_{l}}\left(F_{s}^{l}-\mu_{s}^{l}\right)^{2}}.
\end{equation}

The pixel-wise feature embedding values, known as texture information \cite{stylecontent2}, represent the detailed local content of the scene, and are captured in the intermediate feature map \(F_s^l\) of an encoder \(\phi_{source}\).
The pixel-wise texture value can be formulated as follows:
\begin{equation}
\textstyle
F_s^l(u,v)\in \mathbb{R}^d.
\end{equation}
The layer notation \(l\) has been omitted to simplify the explanation of the equations introduced below.

\subsection{Overview}
\label{Overview}
The overall framework of ST-Seg is illustrated in Fig. \ref{mainfig}. SE method guides our network to accurately predict ground truth labels, even when the source domain's style is substituted with a new one, ensuring robust performance in distribution-shifted target domains.
To create a variety of new styles absent in the source domain, the style generator \(g(\cdot)\) learns a realistic distribution of styles from unlabeled ImageNet data \(X_r \in \mathbb{R}^{H\times W\times 3} \), and the style sampler \(s(\cdot)\) extracts diverse styles in an unbiased manner. These newly generated styles \(\mu_g\) and \(\sigma_g\) are then substituted into the source features \(F_s\), resulting in style-augmented features \(F_{sa}\). These features propagate through the network, producing style-augmented predictions \(P_{sa}\), which are compared with the ground truth \(Y_s\) to compute the style expansion loss \(\mathcal{L}_{style}\). The network parameters learned from these style-augmented predictions are shared with the source network \(\phi_{source}\), enabling the network to adapt to a wide range of distribution-shifted target domains.

The TR method is inspired by the fact that most materials in off-road environments consist of natural elements.
We pre-train a deep texture manifold \(\phi_{texture}\) that effectively discriminates the texture information of these natural elements and regulates the texture information shared from the style-augmented predictions by constraining it with the feature distance \(FD\). 

\subsection{Style Expansion}

To overcome the limitations of previous research that relied on removing \cite{sansaw} or randomizing \cite{domainrandom1} style information, we propose a style expansion method that produces realistic and unbiased styles by expanding the distribution of style in a realistic and diverse manner.

\noindent\textbf{Realistic Style Extraction} 
Inspired by the demonstrated effectiveness of utilizing ImageNet to borrow various feature statistics \cite{threeways}, we extract realistic features by feeding ImageNet samples \(X_{r}\) into the encoder. 
In this process, to mitigate the instability of feature extraction during training and to capture significant data trends, we employ the exponential moving average (EMA) of the source encoder \(\phi_{source}\) for style statistics.
Using Eqs. (\ref{statistics1}) and (\ref{statistics2}), we can calculate the channel-wise realistic style statistics \(\mu_{r}^{i}\) and \(\sigma_{r}^{i}\) of the feature \(F_{r}^{i} = [\phi_{source}^{EMA}]_i(X_{r}^i)\) at the current training step \(i\).

If we simply replace the source style \(\mu_s, \sigma_s\) with the extracted realistic style \(\mu_r, \sigma_r\), the expanded style distribution will be heavily affected by the randomness of sampling \(X_{r}^i\) from \(X_r\), as the new style is determined by which samples are extracted from ImageNet.
Consequently, this can lead to performance fluctuations with each training seed and restrict the style distribution, as the estimated distribution relies on randomness and is constrained by ImageNet features.
To address this issue, we designed a style-generator \(\bold{g}(\cdot)\) and a style-sampler \(\bold{s}(\cdot)\) to learn a realistic style distribution model without relying on ImageNet data, ensuring a balanced experience of both common and rare styles.

\noindent\textbf{Style-generator}
Generally, deep learning features follow a Gaussian distribution \cite{deepgaussian}. Given the IID assumption of ImageNet samples \(X_r\), the feature \(F_r^i\) obtained in each iteration follows an independent Gaussian distribution relative to features from other samples \(F_r^j\), as
\begin{equation}
F_{r}^{i} \sim \mathcal{N}(\mu_{r}^{i}, (\sigma_{r}^{i})^2), \; F_{r}^{i}\,\bot\,F_{r}^{j}\; for\; all\; i\neq j .
\end{equation}
This indicates that each feature can be modeled as an independent Gaussian distribution, uncorrelated with features from other samples. Since the means of multiple independent Gaussian distributions follow a Gaussian distribution and their standard deviation follow a Gamma distribution \cite{statistical}, the realistic style \(\mu_r\), \(\sigma_r\) can be formulated as follows:
\begin{equation}
\{\mu_r\}_{i=1}^{n} \sim \mathcal{N}(\mu_{\mu}, \sigma_{\mu}^2), \quad \{\sigma_r\}_{i=1}^n \sim \Gamma(k, \theta) .
\end{equation}
We temporarily store the extracted realistic style at each iteration \(i\) in a style-buffer. 
Using the accumulated \(\mu_r\) and \(\sigma_r\) as observed samples, we update the parameters of each distribution \(\mu_\mu, \sigma_\mu, k, \theta\) with Bayesian update \cite{beyesian} manner whenever a certain number of samples in style-buffer are gathered. The style-generator \(\bold{g}(\cdot)\) formulated as follows:
\begin{equation}
(\mu_\mu^*,\sigma_\mu^*, k^*,\theta^*) \leftarrow \boldsymbol{g}(\{\mu_r\}_{i=1}^n,\{\sigma_r\}_{i=1}^n;\:\mu_\mu,\sigma_\mu,k,\theta).
\end{equation}

   
\noindent \textbf{Style-sampler}
To address the issue of randomness, we designed the style-sampler \(\bold{s}(\cdot)\) to sample in a balanced way across all ranges using stratified sampling \cite{stratified},
    a variance reduction method that divides the population  \(Q\) into distinct subgroups and samples from each to accurately reflect the original standard deviation.
Specifically, in each training batch, subgroups are predefined based on the number of samples in the batch, and generated samples are evenly assigned to these subgroups. 
We define each subgroup boundaries \(s_b^\mu, s_b^\sigma\) for the generated mean and standard deviation distribution as:
\begin{equation}
s_b^{\mu}= \mu_\mu^* + \sigma_\mu^* \cdot \mathcal{N}^{-1}(b/B),\hspace{4pt} s_b^{\sigma}= \Gamma^{-1}(b/B;k^*, \theta^*)
\end{equation}
where \(b\) is subgroup index, \(B\) represents the number of subgroups, which equals the training batch size, and \(\mathcal{N}^{-1}, \Gamma^{-1}\) are the inverse CDFs of the Gaussian and Gamma distributions, respectively.
Using the computed subgroup boundaries, each batch at every training step can be assigned unbiased samples of generated style statistics.
 
\begin{equation}
(\{\mu_g^b\}_{b=1}^B, \{\sigma_g^b\}_{b=1}^B) \leftarrow \boldsymbol{s}(\mu_\mu^*,\sigma_\mu^*, k^*,\theta^*)
\end{equation}
Through the above derivation, the style-sampler \(\bold{s}(\cdot)\) samples diverse samples, effectively balancing the representation of styles and ensuring access to both common and rare styles within the modeled distribution.

\noindent \textbf{Style Substitution}
Now, we can simply substitute the style of the source feature \(F_s\) with output of the style-sampler for each train-batch. 
\begin{equation}
F_{sa}=\sigma_{g}\cdot\frac{F_{s}-\mu_{s}}{\sigma_{s}}+\mu_{g}
\end{equation}
Note that, the distribution of \(F_{s}\) is re-normalized based on the generated statistics, which means that the spatial information is preserved in \(F_{sa}\). 
The \(l\)-th style-augmented feature \(F_{sa}^{l}\) is then fed into next layer \(l+1\). 
In this process, the generated style information propagates to the subsequent layers.

\noindent \textbf{Style Expansion Loss}
We predict two class probability maps: one from the source path, where the source feature flows, and another from the additional path, where the style-augmented feature flows.
The segmentation network is trained to predict the ground truth even when style augmentation is applied to the data. To achieve this, the cross-entropy loss is computed between the style-augmented probability map and the ground truth Y as follows:
\begin{equation}
\textstyle
\mathcal{L}_{style}=-\frac{1}{HW}\sum_{h=1}^{H}\sum_{w=1}^{W}\sum_{c=1}^{C}Y_{s}\log{P_{sa}}
\end{equation}
Style loss encourages the network to learn style-invariant feature representations by introducing controlled perturbations in the feature space. It acts as a regularization mechanism, reducing overfitting to specific styles and enhancing robustness to unseen style variations.
To ensure consistency between the source prediction \(P_s\) and the style-augmented prediction \(P_{sa}\), we use the Kullback-Leibler (KL) divergence loss as the align loss:
\begin{equation}
\textstyle
\mathcal{L}_{align}=-\frac{1}{HW}\sum_{h=1}^{H}\sum_{w=1}^{W}\sum_{c=1}^{C}P_{s}\log\frac{P_{s}}{P_{sa}}
\end{equation}

\subsection{Texture Regularization}

Since style information affects the global scene, it also affects local textures that constitute it \cite{stylecontentorthogonal}. Thus, altering the style can lead to changes in the texture as well.
In off-road environments, which primarily consist of natural elements, textures exhibit limited variation, even though they appear in diverse forms across different locations and times.
Regularizing the texture information of these natural elements stabilizes feature representation, mitigating the instability caused by fluctuating texture information.

\noindent \textbf{Deep Texture Manifold}
We hypothesize that if an auxiliary encoder effectively extracts the deep texture manifold, it can guide the model in preserving texture knowledge, similar to the knowledge distillation method \cite{distillation}.
To achieve this, we pre-trained a texture encoder, \(\phi_{texture}\), using a patch-level classification task by sampling natural materials from the GTOS-mobile dataset \cite{gtos}, which is publicly released in a study focused on the deep texture manifold.
This frozen pre-trained texture encoder can be utilized as the teacher network during the training phase by extracting the deep texture manifold of a training sample, formulated as \(F_{t}=\phi_{texture}(X_s)\), where \(F_{t}\in \mathbb{R}^{d\times H\times W}\).

\noindent \textbf{Feature Distance}
We designed a regularization term at the latent feature level to train the segmentation encoder to extract features that approximate those of the deep texture manifold.
Since latent features are not probabilities like softmax outputs, we use L2 distance instead of cross-entropy, differing from traditional knowledge distillation:
\begin{equation}
FD(i,j) = \lVert F_t^l(i,j) - F_s^l(i,j) \rVert_2
\end{equation}
where, \(i\) and \(j\) are the pixel coordinates.
However, enforcing all pixels to approximate the deep texture manifold could compromise semantic information. To address this, we selectively applied regularization to the natural element class \(\mathcal{C}_{natural}\) (e.g., vegetation, soil, rock) by calculating the feature distance only for regions defined by the binary mask \(M_{t}\), which corresponds to the natural element class in the ground-truth labels:
\begin{equation}
M_{t}=\sum_{c=1}^{C}Y_{s}\cdot[c\in\mathcal{C}_{natural}].
\end{equation}

\noindent \textbf{Texture Regularization Loss} We designed the texture regularization loss to selectively target the texture-only regions using a binary mask, defined as:
\begin{equation}
\mathcal{L}_{tex}=\frac{\sum_{h=1}^{H^l}\sum_{w=1}^{W^l}\gamma^{l}FD(i,j)\cdot M_{t}}{\sum_{h=1}^{H^l}\sum_{w=1}^{W^l}M_{t}}
\end{equation}
where \(\gamma^l\) is layer-wise weight, inspired by previous deep learning studies suggesting that local texture information is predominantly processed in shallow layers \cite{textureshallow}. Detailed information about the hyperparameters is provided in subsection \ref{implementation}.

\subsection{Training and Inference}

\noindent\textbf{Training phase}
As illustrated in Fig. \ref{mainfig}, the training phase optimizes the source encoder and decoder to minimize the proposed style loss, align loss, texture loss and the standard cross-entropy loss \(\mathcal{L}_{ce}\) \cite{celoss}.
\begin{equation}
\mathcal{L} = \mathcal{L}_{ce} + \mathcal{L}_{style} + \mathcal{L}_{align} + \mathcal{L}_{tex}
\end{equation}

\noindent\textbf{Inference phase}
During training, the encoder and decoder are exposed to a variety of realistic styles through additional data and modules. 
Through the trained encoder-decoder, robust predictions can be made in real-time without the need for additional components or computations.
\begin{table*}[t]
\renewcommand{\arraystretch}{1.4}

\caption{Performance Comparisons for Internal Distribution-shifted Target Data (RGR-C).}
\begin{adjustbox}{width=\linewidth,center}
\large
\begin{tabular}{l|c|cccccccc|c}

\midrule 
Method (mIoU / mAcc) & Memory / Latency & Brightness & Contrast & Defocus-blur & Motion-blur & Impulse-noise & Gaussian-noise & Snow-noise & Frost-lens & \textit{Avg.} \\
\midrule
BiseNetv2 \cite{bisenetv2} & 14.78 M / 108.7 ms & 71.56 / 80.88 & 38.78 / 50.63 & 53.90 / 67.68 & 64.39 / 77.56 & 24.94 / 37.66 & 26.87 / 41.21 & 32.82 / 43.16 & 38.72 / 47.67 & 43.99 / 55.80 \\
DeepLabv3+ \cite{deeplabv3+} & 14.47 M / 149.3 ms & 77.55 / 86.01 & 48.56 / 62.86 & 52.80 / 66.30 & 66.48 / 80.06 & 26.66 / 40.43 & 29.07 / 42.80 & 37.58 / 48.98 & 45.42 / 57.01 & 48.02 / 60.56 \\
MobileNetv3 \cite{mobilenet} & \textbf{3.28 M} / \underline{84 ms} & 73.68 / 83.48 & 47.73 / 62.67 & 51.12 / 71.38 & 64.36 / 78.54 & 33.59 / 45.74 & 35.10 / 47.26 & 38.70 / 51.92 & 44.63 / 57.11 & 48.61 / 62.26 \\
SegFormer \cite{segformer} & \underline{3.72 M} / \textbf{74.6 ms} & \underline{80.13} / \textbf{88.27} & \underline{67.04} / \underline{79.06} & 60.75 / 70.93 & 72.79 / 82.89 & 36.28 / 50.63 & 39.23 / 52.46 & \underline{43.39} / \underline{60.57} & 50.92 / \underline{62.03} & \underline{56.32} / \underline{68.36} \\
GaNav \cite{ganav} & 6.10 M / 102 ms & 76.19 / 85.09 & 55.18 / 65.97 & \underline{69.56} / \underline{80.44} & \underline{74.37} / \underline{84.18} & \underline{42.39} / \underline{53.08} & \underline{45.75} / 57.53 & 38.48 / 49.56 & 42.11 / 52.08 & 55.50 / 65.99 \\
Lin et al. \cite{sansaw} & 11.00 M / 112.4 ms & 79.08 / 87.84 & 62.79 / 75.91 & 58.22 / 69.77 & 69.32 / 81.09 & 36.79 / 51.79 & 44.82 / \underline{57.75} & 42.41 / 57.55 & \underline{51.68} / 54.32 & 55.64 / 68.25 \\

ST-Seg (Ours) & \underline{3.72 M} / \textbf{74.6 ms} & \textbf{80.29} / \underline{88.01} & \textbf{70.46} / \textbf{81.51} & \textbf{76.45} / \textbf{85.34} & \textbf{78.79} / \textbf{87.06} & \textbf{62.26} / \textbf{70.60} & \textbf{61.22} / \textbf{69.50} & \textbf{67.44} / \textbf{76.81} & \textbf{59.87} / \textbf{70.24} & \textbf{69.60} / \textbf{78.63} \\
\bottomrule
\end{tabular}
\end{adjustbox}

\vspace{-0cm}
\label{corrupt_table}
\end{table*}

\begin{figure*}[thpb]
      \centering
      \includegraphics[width=0.9\linewidth]{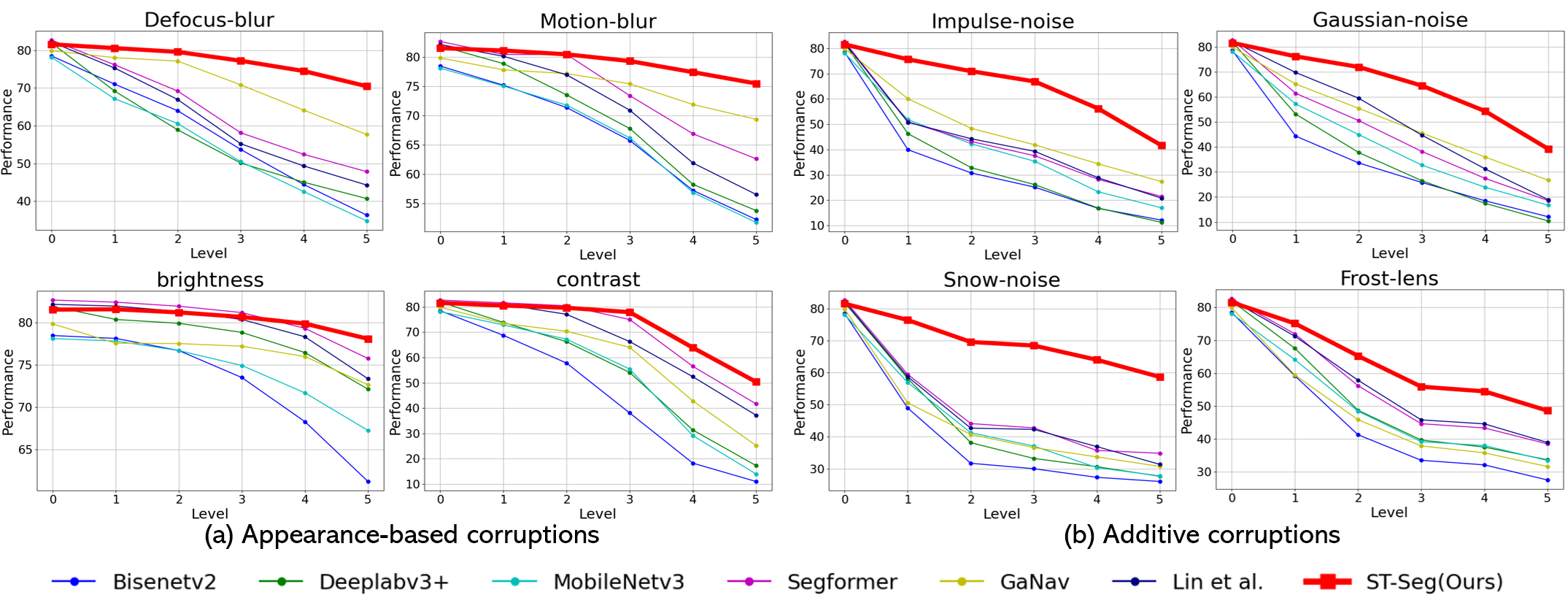}
      \captionsetup{font=small}
      \caption{\textbf{Results and Comparisons across Corruption Levels for Internal Distribution Shift.} We illustrated the performance trends for each type of corruption using line charts. In each chart, the x-axis represents the corruption level, while the y-axis indicates performance. The bold red line represents the performance of our proposed ST-Seg, which demonstrates the best ability to maintain its original performance even as the corruption levels increase.}
      \vspace{-0.5cm}
      \label{corrupt level}
   \end{figure*}

\section{Results and Analysis}
\subsection{Experimental Setup}

We designed experiments to validate the effectiveness of ST-Seg in addressing the internal and external distribution shifts introduced earlier.
To simulate internal distribution shifts, which arise from sensor corruption during navigation in unstructured environments, we adopted well-established image corruption methods \cite{benchmarking} to construct the validation set.
For external distribution shifts, which occur due to discrepancies between the source and target domains encountered during the operational phase, we split the training and validation data at the dataset level to evaluate performance.
We benchmark our models on the standard segmentation metrics: mean intersection over Union (mIoU), which evaluates the overlap between predicted and ground truth regions, and mean pixel accuracy (mAcc), which measures the average proportion of correctly classified pixels per class.
\begin{equation}
\textstyle
    IoU_c = \frac{\sum_{x,y} \mathds{1}(P(x,y)=c \; \text{and} \; Y(x,y)=c)}{\sum{x,y} \mathds{1}(P(x,y)=c \;  \text{or} \; Y(x,y)=c)}, \hspace{5pt} mIoU= \frac{\sum_{c} IoU_c}{\sum_{C}1}
\end{equation}
\begin{equation}
\textstyle
mAcc = \frac{\sum_{c\in C} (\sum_{x,y,Y(x,y)=c} \mathds{1}(P(x,y)=Y(x,y)))}{\sum_C 1}
\end{equation}
For an image \(I\), let \(P(x,y)\) and \(Y(x,y)\) be the prediction and ground truth labels at pixel \((x,y)\), \(\mathds{1}(\cdot)\) be the indicator function, \(c\) be the class index, and \(C\) as the set of class labels.

We also qualitatively validated ST-Seg on challenging real-world data collected from both our Clearpath Husky UGV and the small mobile UGV, Frodobots \cite{frodobots_lab_2024}, demonstrating its effectiveness in highly unstructured environments.
The latency of the models used for performance comparison was measured on the NVIDIA Jetson AGX Orin.

\subsection{Implementation Details}
\label{implementation}
\noindent \textbf{Backbone}
The backbone architecture of ST-Seg is the smallest version of the mixed vision transformer (MiT-B0) \cite{segformer}, which efficiently extracts multi-scale features, ensuring fast inference while achieving notable performance in semantic segmentation benchmarks.

\noindent \textbf{Off-road Dataset}
RUGD \cite{rugd}, RELLIS \cite{rellis}, and GOOSE \cite{goose} datasets offer detailed class categorizations and high-quality labels, while the TAS \cite{tas}, DeepScene \cite{deepscene}, and YCOR \cite{ycor} datasets provide simpler categorizations with lower-quality labels. Thus, the first three datasets were combined to form the source training domain, termed RGR.
To validate internal distribution shifts, we generated RGR-C as target domain by applying image corruption methods to RGR. The applied corruptions include brightness, contrast, defocus blur, impulse noise, Gaussian noise, snow noise, and frost lens effects.
For external distribution shift validation, RGR used as the source domain, while TDY served as the target domain.
Official splits were employed for all datasets.
To standardize class categories across different datasets, we remapped based on traversability levels relevant to off-road navigation \{background, smooth, rough, bumpy, soft vegetation, hard vegetation, puddle, and obstacle.\}.


\noindent \textbf{Details of hyper-parameter setting}
All models were trained using the hyperparameters specified by their respective authors, with a batch size of 8.  
Optimization was performed using the AdamW optimizer with a learning rate of 0.00006, betas (0.9, 0.999), and a weight decay of 0.01.  
A polynomial learning rate policy with a power of 1.0 was applied.  
The EMA model was updated with a smoothing factor of $\alpha = 0.7$, and the style-sampler's population size $Q$ was set to 10,000.  
Style-augmented features through the SE method were applied only to the first two layers. 
The texture regularization loss weights $\gamma^l$ were set to \{0.05, 0.025, 0.01, 0.005\} for each layer $l$.  
Our implementation is based on MMSeg \cite{mmseg}. All comparison models were initialized with ImageNet pre-trained weights, and performance was averaged over five training seeds.

\subsection{Results and Compraisons}
\noindent \textbf{Baselines}
To ensure real-time operation on the mobile robot, lightweight segmentation models with strong performance were used as baselines \cite{bisenetv2, deeplabv3+, mobilenet, segformer}. These models represent the state of the art in real-time semantic segmentation.
Additionally, GaNav \cite{ganav}, which demonstrated the best performance in off-road environments, and the method proposed in \cite{sansaw}, which applies instance normalization to extract general information, were also used as baselines.

\begin{figure}[t]
      \centering
      \includegraphics[width=1\linewidth]{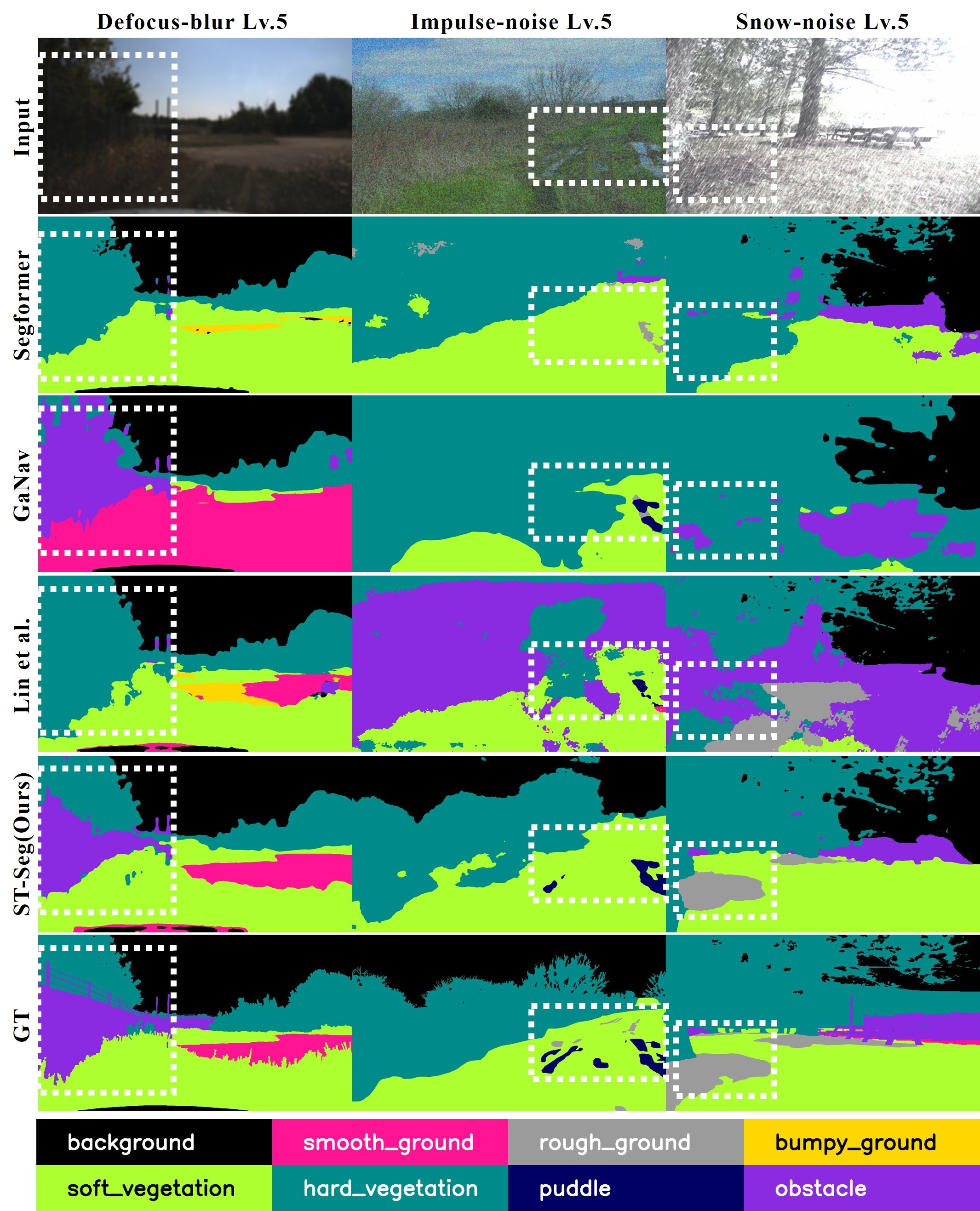}
      \captionsetup{font=small}
      \caption{\textbf{Qualitative results on RGR-C.} Our method achieves better segment of different pixel-wise classes, as shown in the white box, closely matching the ground truth.}
      \vspace{-0.4cm}
      \label{expfig}
   \end{figure}

\noindent \textbf{Results on Internal Distribution Shift}
To verify the robustness in internal distribution shift, we conducted performance validation on RGR-C and presented in TABLE \ref{corrupt_table}.
All models were trained on the training split of the RGR dataset and evaluated on the corrupted data from the RGR-C validation split.
With our proposed learning strategies, ST-Seg achieved an average mIoU of 69.60\% across all types and levels of corruption, 
showing a +13.28\% improvement over the second-best baseline \cite{segformer}.
The main improvements, observed for defocus-blur (+15.7\%), impulse-noise (+25.98\%), gaussian-noise (+21.99\%), and snow-noise (+24.05\%), demonstrate the effectiveness of our model's ability to learn realistic styles from ImageNet, enabling robust performance on corruption types that closely mirror real-world scenarios.
Performance trends across corruption levels are shown in Fig. \ref{corrupt level}, highlighting the ability to maintain performance of ST-Seg as corruption level increases.
Qualitative results in Fig. \ref{expfig} demonstrate that ST-Seg produces robust predictions, closely matching the ground truth even on corrupted data.

\begin{table}[h!]
\renewcommand{\arraystretch}{1.2}
\caption{Performance Comparisons for External Distribution-shifted Target Data. (TDY)}
\begin{adjustbox}{width=\linewidth}
\begin{tabular}{l|c|ccc}
\toprule
Method (mIoU / mAcc) & RGR & TAS & Deepscene & YCOR   \\ 
\midrule
BiseNetv2 \cite{bisenetv2}& 78.48 / 86.32&43.38 / 59.28 & 34.60 / 36.80 & 35.97 / 47.82 \\ 
DeepLabv3+ \cite{deeplabv3+}& 81.93 / \underline{89.38} & \underline{53.59} / \underline{68.40} & 48.04 / 56.23 & 36.89 / 45.86 \\ 
MobileNetv3 \cite{mobilenet}& 78.12 / 86.44 & 50.02 / 63.41 & 40.08 / 43.69& 31.07 / 45.38 \\
SegFormer \cite{segformer}& \textbf{82.68} / \textbf{89.56}  & 39.99 / 55.62 & 40.10 / 46.05 & 34.39 / \underline{48.11} \\ 
GaNav \cite{ganav}& 79.86 / 87.87 & 45.80 / 62.89 & 52.62 / 58.62 & 34.57 / 47.52 \\ 
Lin et al. \cite{sansaw} & \underline{82.17} / 89.21 & 50.62 / 65.76 & \underline{54.78} / \underline{62.61} & \underline{36.99} / 48.06 \\
ST-Seg (Ours) & 81.55 / 88.64 & \textbf{56.22} / \textbf{69.37} & \textbf{59.96} / \textbf{67.43} & \textbf{40.13} / \textbf{50.57} \\
\bottomrule
\end{tabular}
\end{adjustbox}
\vspace{-0.7cm}
\label{TDY}
\end{table}
   
\noindent \textbf{Results on External Distribution Shift}
To verify robustness under external distribution shift, we conducted experiments on unseen target data and presented in TABLE \ref{TDY}.
All models were trained on the training split of the RGR and evaluated on both the validation split of the source domain (RGR) and various unseen target domains (TDY).

Our ST-Seg model shows an improvement in mIoU compared to the SegFormer \cite{segformer} with the same backbone: TAS (+16.23\%), Deepscene (+19.86\%), and YCOR (+5.74\%), despite slight performance decrease in the source domain (-1.13\%).
These results indicate that the slight decline in the source domain RGR is a reasonable trade-off, given the significant performance gains 
achieved across various external distribution shifted target domains (TDY), including internal distribution shifted target domain (RGR-C).

\noindent \textbf{Navigation Safety and Efficiency.} To validate safety and efficiency from a navigation perspective, we conducted experiments on precision and recall on RGR-C. Precision, which measures the proportion of correctly predicted positives, indicates that high precision for the traversable class reduces the risk of misclassifying non-traversable areas as traversable, thereby lowering collision risks. Conversely, high precision for the non-traversable class means fewer occurrences of misclassifying traversable areas as non-traversable, which contributes to generating more efficient paths. Recall, the proportion of correctly predicted positives out of all actual positives, suggests that high recall for the traversable class reflects high navigation efficiency, similar to precision. On the other hand, recall for the non-traversable class is directly related to navigation safety.
\begin{equation}
\textstyle
    precision=\frac{TP}{TP+FP}, \hspace{10pt} recall=\frac{TP}{TP+FN}
\end{equation}
As shown in Fig. 6, our ST-Seg outperforms the baseline model \cite{segformer} in all cases, highlighting its advantages in both driving safety and efficiency.

\noindent \textbf{Effectiveness of SE, TR} 
We conducted an ablation study to evaluate our approach on the corrupted seen domain RGR-C and the unseen domain TDY, as shown in TABLE \ref{ablation}.
As a baseline, we used SegFormer \cite{segformer} with the same backbone network.
To demonstrate the advantage of using realistic style statistics, we first experimented with two approaches: applying random style adjustment \cite{domainrandom1} to the baseline and training with a naive substitution of realistic style statistics from ImageNet.

\begin{figure}[t]
      \centering
      \includegraphics[width=1\linewidth]{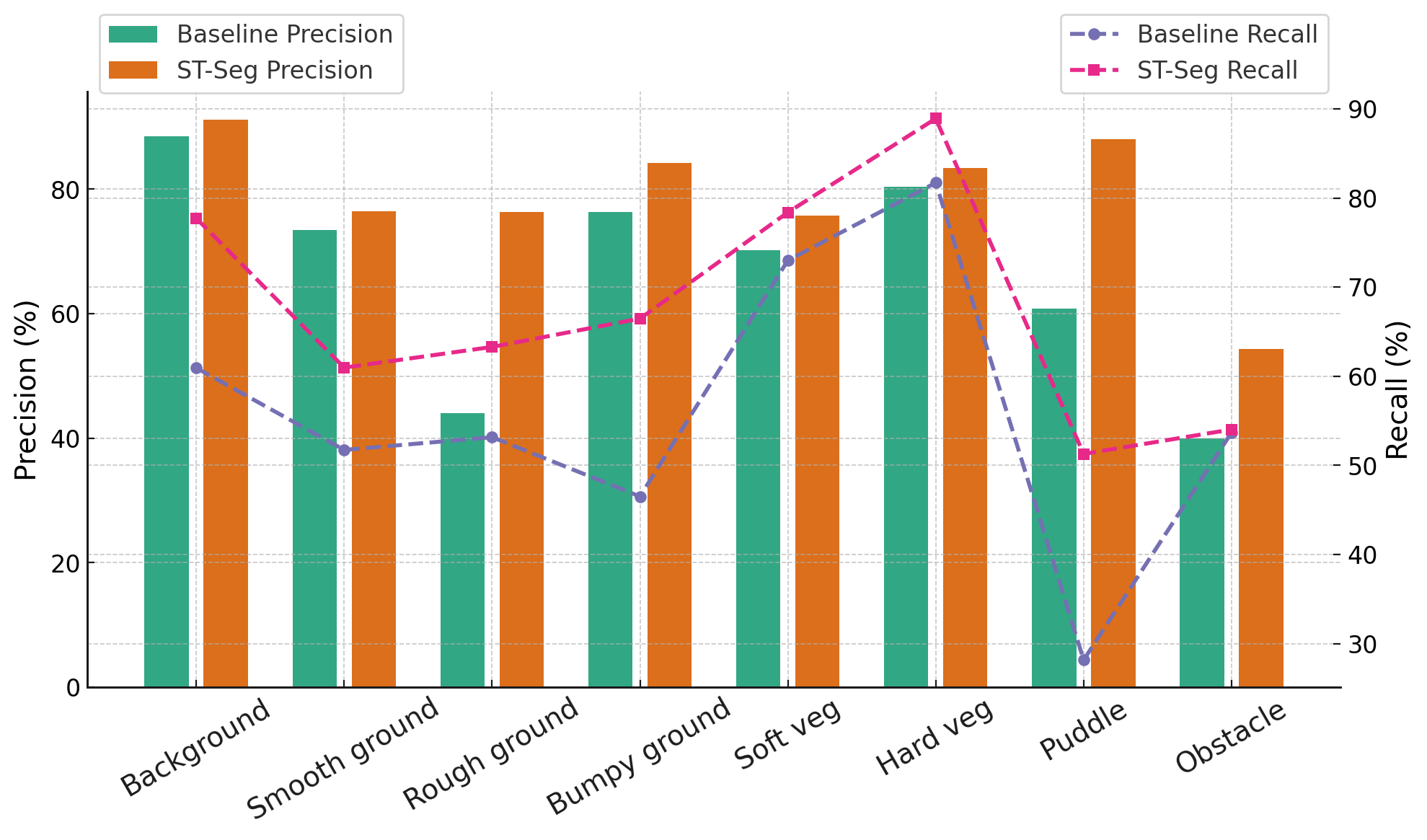}
      \captionsetup{font=small}
      \caption{\textbf{Precision and Recall Comparisons on RGR-C.}} 
      \vspace{-0.1cm}
      \label{precision}
   \end{figure}

Each approach showed an increase in average mIoU of +5.68\% and +7.65\%, respectively, compared to the baseline.
The SE method, which utilizes style-generator and style-sampler for realistic and unbiased style generation, demonstrated a +2.46\% improvement compared to the naive realistic style approach. 

\begin{table}[h]
\renewcommand{\arraystretch}{1.2}
\caption{Ablation Study of Proposed Method.}
\begin{adjustbox}{width=\linewidth}
\begin{tabular}{l|cc|c}
\toprule
Method (mIoU / mAcc) & RGR-C & TDY & Avg. \\
\midrule
Baseline \cite{segformer} & 56.32 / 68.36  & 38.16 / 49.93 & 47.24 / 59.15 \\
Baseline + Rand style \cite{domainrandom1} & 62.20 / 71.90 & 43.63 / 53.41 & 52.92 / 62.66 \\
Baseline + Real style & 64.12 / 74.37 & 45.66 / 56.39 & 54.89 / 65.38\\
Baseline + SE & 66.67 / 75.87 & 48.02 / 58.35 & 57.35 / 67.11 \\
Baseline + TR& 58.78 / 69.31 & 46.70 / 58.59 & 52.24 / 63.95 \\
Baseline + SE + TR & \textbf{69.60} / \textbf{78.63} & \textbf{52.10} / \textbf{62.46} & \textbf{60.85} / 70.55 \\
\bottomrule
\end{tabular}
\end{adjustbox}
\vspace{-0.4cm}
\label{ablation}
\end{table}

Furthermore, when combined with the TR method to compensate for texture discrimination power, it showed an overall increase of +5.96\%.
This demonstrates that the texture loss \(\mathcal{L}_tex\) from the TR texture feature \(F_t\)  contributes to consistent texture representation, mitigating the instability of local textures caused by the SE module.


\subsection{Real world evaluation with mobile robot}
\begin{figure}[t]
      \centering
      \includegraphics[width=0.93\linewidth]{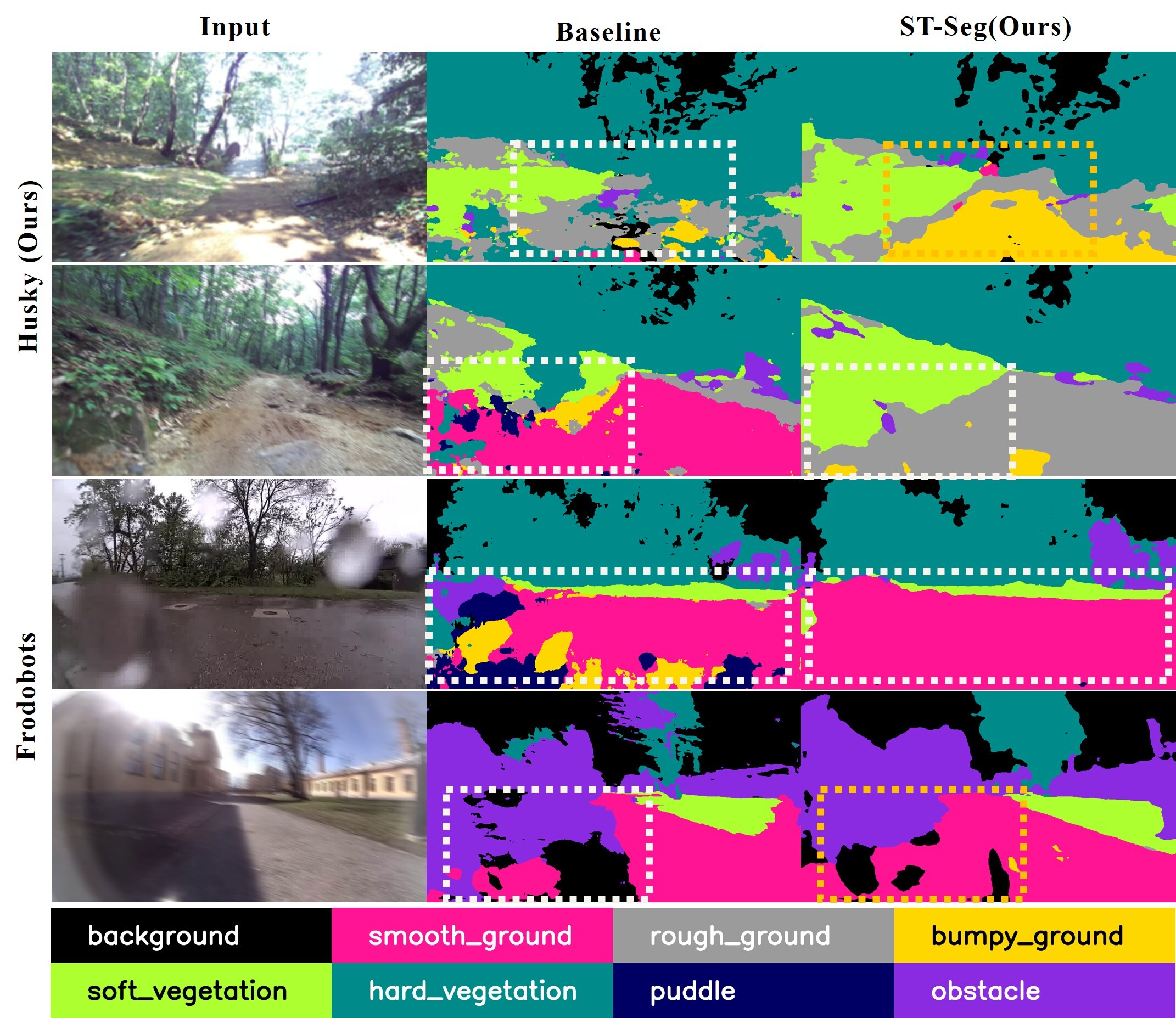}
      \captionsetup{font=small}
      \caption{\textbf{Qualitative results on challenging real-world data.} Compared to the baselines, which produce noisy and incorrect predictions, our method achieves clearer and more accurate segmentation, as shown in the white box. The areas highlighted with yellow boxes show that, in cases of severe corruption and extreme edge cases, the predictions are still not as perfect as humans.}
      \vspace{-0.5cm}
      \label{realfig}
   \end{figure}

We also qualitatively validated the proposed learning framework in challenging real-world scenes using our Clearpath Husky and Frodobots \cite{frodobots_lab_2024}. With the Husky, experiments were carried out on rough terrain under low-light mountain conditions. For the Frodobots, we experiments under challenging factors such as debris on the lens and intense backlighting from sunlight. As shown in Fig. \ref{realfig}, ST-Seg produces more robust predictions than the baseline \cite{segformer} on distribution-shifted target data. It not only delivers more accurate semantic predictions but also significantly reduces segmentation noise, resulting in more consistent outputs across surrounding areas.

\section{CONCLUSIONS}

In this paper, we introduced ST-Seg, a novel learning framework to enhance the performance of semantic segmentation models in off-road environments under distribution shifts. By expanding the limited styles of the source domain through SE and stabilizing feature representation via TR, we achieved significant improvements in model robustness.
Our experiments demonstrated that ST-Seg maintains performance effectively under both internal and external distribution-shifted target domains, enhancing the real-world applicability of off-road semantic segmentation.
Future work will focus on enabling the model to incrementally adapt to distribution-shifted target domains during operation, making the navigation framework more flexible and responsive to real-time environmental changes.








\bibliographystyle{ieeetr}
\bibliography{ref}

@article{hao2020brief,
  title={A brief survey on semantic segmentation with deep learning},
  author={Hao, Shijie and Zhou, Yuan and Guo, Yanrong},
  journal={Neurocomputing},
  volume={406},
  pages={302--321},
  year={2020},
  publisher={Elsevier}
}

@article{segformer,
  title={SegFormer: Simple and efficient design for semantic segmentation with transformers},
  author={Xie, Enze and Wang, Wenhai and Yu, Zhiding and Anandkumar, Anima and Alvarez, Jose M and Luo, Ping},
  journal={Advances in neural information processing systems},
  volume={34},
  pages={12077--12090},
  year={2021}
}

@inproceedings{rugd,
  author = {Wigness, Maggie and Eum, Sungmin and Rogers, John G and Han, David and Kwon, Heesung},
  title = {A RUGD Dataset for Autonomous Navigation and Visual Perception in Unstructured Outdoor Environments},
  booktitle = {International Conference on Intelligent Robots and Systems (IROS)},
  year = {2019}
}

@misc{rellis,
      title={RELLIS-3D Dataset: Data, Benchmarks and Analysis}, 
      author={Peng Jiang and Philip Osteen and Maggie Wigness and Srikanth Saripalli},
      year={2020},
      eprint={2011.12954},
      archivePrefix={arXiv},
      primaryClass={cs.CV}
}

@article{goose,
    author = {Peter Mortimer and Raphael Hagmanns and Miguel Granero
              and Thorsten Luettel and Janko Petereit and Hans-Joachim Wuensche},
    title = {The GOOSE Dataset for Perception in Unstructured Environments},
    url={https://arxiv.org/abs/2310.16788},
    conference={2024 IEEE International Conference on Robotics and Automation (ICRA)},
    year = {2024}
}

@InProceedings{tas,
  author    = {Kai A. Metzger AND Peter Mortimer AND Hans-Joachim Wuensche},
  title     = {{A Fine-Grained Dataset and its Efficient Semantic Segmentation for Unstructured Driving Scenarios}},
  booktitle = {International Conference on Pattern Recognition (ICPR2020)},
  year      = {2021},
  address   = {Milano, Italy (Virtual Conference)},
  month     = jan,
}

@InProceedings{deepscene,
author = {Abhinav Valada and Gabriel Oliveira and Thomas Brox and Wolfram Burgard},
title = {Deep Multispectral Semantic Scene Understanding of Forested Environments using Multimodal Fusion},
booktitle = {International Symposium on Experimental Robotics (ISER)},
year = {2016},
}

@inproceedings{ycor,
  title={Real-time semantic mapping for autonomous off-road navigation},
  author={Maturana, Daniel and Chou, Po-Wei and Uenoyama, Masashi and Scherer, Sebastian},
  booktitle={Field and Service Robotics},
  pages={335--350},
  year={2018},
  organization={Springer}
}

@Article{agriculture,
AUTHOR = {Oliveira, Luiz F. P. and Moreira, António P. and Silva, Manuel F.},
TITLE = {Advances in Agriculture Robotics: A State-of-the-Art Review and Challenges Ahead},
JOURNAL = {Robotics},
VOLUME = {10},
YEAR = {2021},
NUMBER = {2},
ARTICLE-NUMBER = {52},
URL = {https://www.mdpi.com/2218-6581/10/2/52},
ISSN = {2218-6581},
DOI = {10.3390/robotics10020052}
}

@Inbook{disaster,
author="Murphy, Robin R.
and Tadokoro, Satoshi
and Kleiner, Alexander",
editor="Siciliano, Bruno
and Khatib, Oussama",
title="Disaster Robotics",
bookTitle="Springer Handbook of Robotics",
year="2016",
publisher="Springer International Publishing",
address="Cham",
pages="1577--1604",
isbn="978-3-319-32552-1",
doi="10.1007/978-3-319-32552-1_60",
url="https://doi.org/10.1007/978-3-319-32552-1_60"
}

@ARTICLE{arte,
  author={Yoon, Hyung-Suk and Hwang, Ji-Hoon and Kim, Chan and Son, E In and Yoo, Se-Wook and Seo, Seung-Woo},
  journal={IEEE Robotics and Automation Letters}, 
  title={Adaptive Robot Traversability Estimation Based on Self-Supervised Online Continual Learning in Unstructured Environments}, 
  year={2024},
  volume={9},
  number={6},
  pages={4902-4909},
  doi={10.1109/LRA.2024.3386451}}

@article{stratified,
  title={Stratified sampling},
  author={Singh, Ravindra and Mangat, Naurang Singh and Singh, Ravindra and Mangat, Naurang Singh},
  journal={Elements of survey sampling},
  pages={102--144},
  year={1996},
  publisher={Springer}
}

@book{statistical,
  title={Statistical Inference},
  author={Casella, George and Berger, Roger L.},
  year={2002},
  edition={2nd},
  publisher={Duxbury Press},
  address={Pacific Grove, CA},
}

@inproceedings{offseg,
  title={Offseg: A semantic segmentation framework for off-road driving},
  author={Viswanath, Kasi and Singh, Kartikeya and Jiang, Peng and Sujit, PB and Saripalli, Srikanth},
  booktitle={2021 IEEE 17th international conference on automation science and engineering (CASE)},
  pages={354--359},
  year={2021},
  organization={IEEE}
}

@article{uncertainty,
  title={Uncertainty-aware perception models for off-road autonomous unmanned ground vehicles},
  author={Yang, Zhaoyuan and Tan, Yewteck and Sen, Shiraj and Reimann, Johan and Karigiannis, John and Yousefhussien, Mohammed and Virani, Nurali},
  journal={arXiv preprint arXiv:2209.11115},
  year={2022}
}

@article{ganav,
  title={Ga-nav: Efficient terrain segmentation for robot navigation in unstructured outdoor environments},
  author={Guan, Tianrui and Kothandaraman, Divya and Chandra, Rohan and Sathyamoorthy, Adarsh Jagan and Weerakoon, Kasun and Manocha, Dinesh},
  journal={IEEE Robotics and Automation Letters},
  volume={7},
  number={3},
  pages={8138--8145},
  year={2022},
  publisher={IEEE}
}

@misc{
batch,
title={Revisiting Batch Normalization For Practical Domain Adaptation},
author={Yanghao Li and Naiyan Wang and Jianping Shi and Jiaying Liu and Xiaodi Hou},
year={2017},
url={https://openreview.net/forum?id=BJuysoFeg}
}

@article{sansaw,
  title={Real-Time Segmentation of Unstructured Environments by Combining Domain Generalization and Attention Mechanisms},
  author={Lin, Nuanchen and Zhao, Wenfeng and Liang, Shenghao and Zhong, Minyue},
  journal={Sensors},
  volume={23},
  number={13},
  pages={6008},
  year={2023},
  publisher={MDPI}
}

@inproceedings{mobilenet,
  title={Searching for mobilenetv3},
  author={Howard, Andrew and Sandler, Mark and Chu, Grace and Chen, Liang-Chieh and Chen, Bo and Tan, Mingxing and Wang, Weijun and Zhu, Yukun and Pang, Ruoming and Vasudevan, Vijay and others},
  booktitle={Proceedings of the IEEE/CVF international conference on computer vision},
  pages={1314--1324},
  year={2019}
}

@inproceedings{deeplabv3+,
  title={Encoder-decoder with atrous separable convolution for semantic image segmentation},
  author={Chen, Liang-Chieh and Zhu, Yukun and Papandreou, George and Schroff, Florian and Adam, Hartwig},
  booktitle={Proceedings of the European conference on computer vision (ECCV)},
  pages={801--818},
  year={2018}
}

@article{bisenetv2,
  title={Bisenet v2: Bilateral network with guided aggregation for real-time semantic segmentation},
  author={Yu, Changqian and Gao, Changxin and Wang, Jingbo and Yu, Gang and Shen, Chunhua and Sang, Nong},
  journal={International Journal of Computer Vision},
  volume={129},
  pages={3051--3068},
  year={2021},
  publisher={Springer}
}

@article{stylecontent2,
  title={Texture networks: Feed-forward synthesis of textures and stylized images},
  author={Ulyanov, Dmitry and Lebedev, Vadim and Vedaldi, Andrea and Lempitsky, Victor},
  journal={arXiv preprint arXiv:1603.03417},
  year={2016}
}

@inproceedings{gtos,
  title={Deep texture manifold for ground terrain recognition},
  author={Xue, Jia and Zhang, Hang and Dana, Kristin},
  booktitle={Proceedings of the IEEE Conference on Computer Vision and Pattern Recognition},
  pages={558--567},
  year={2018}
}

@article{stylecontentorthogonal,
  title={Style follows content: On the microgenesis of art perception},
  author={Augustin, M Dorothee and Leder, Helmut and Hutzler, Florian and Carbon, Claus-Christian},
  journal={Acta psychologica},
  volume={128},
  number={1},
  pages={127--138},
  year={2008},
  publisher={Elsevier}
}

@article{benchmarking,
  title={Benchmarking neural network robustness to common corruptions and surface variations},
  author={Hendrycks, Dan and Dietterich, Thomas G},
  journal={arXiv preprint arXiv:1807.01697},
  year={2018}
}

@misc{mmseg,
    title={{MMSegmentation}: OpenMMLab Semantic Segmentation Toolbox and Benchmark},
    author={MMSegmentation Contributors},
    howpublished = {\url{https://github.com/open-mmlab/mmsegmentation}},
    year={2020}
}

@inproceedings{adain,
  title={Arbitrary style transfer in real-time with adaptive instance normalization},
  author={Huang, Xun and Belongie, Serge},
  booktitle={Proceedings of the IEEE international conference on computer vision},
  pages={1501--1510},
  year={2017}
}

@article{beyesian,
  title={An essay towards solving a problem in the doctrine of chances},
  author={Bayes, Thomas},
  journal={Biometrika},
  volume={45},
  number={3-4},
  pages={296--315},
  year={1958}
}

@inproceedings{imagenet,
  title={Imagenet: A large-scale hierarchical image database},
  author={Deng, Jia and Dong, Wei and Socher, Richard and Li, Li-Jia and Li, Kai and Fei-Fei, Li},
  booktitle={2009 IEEE conference on computer vision and pattern recognition},
  pages={248--255},
  year={2009},
  organization={Ieee}
}

@article{deepgaussian,
  title={Deep Gaussian mixture models},
  author={Viroli, Cinzia and McLachlan, Geoffrey J},
  journal={Statistics and Computing},
  volume={29},
  pages={43--51},
  year={2019},
  publisher={Springer}
}

@article{ema,
  title={Mean teachers are better role models: Weight-averaged consistency targets improve semi-supervised deep learning results},
  author={Tarvainen, Antti and Valpola, Harri},
  journal={Advances in neural information processing systems},
  volume={30},
  year={2017}
}

@article{celoss,
author = {Krizhevsky, Alex and Sutskever, Ilya and Hinton, Geoffrey E.},
title = {ImageNet classification with deep convolutional neural networks},
year = {2017},
issue_date = {June 2017},
publisher = {Association for Computing Machinery},
address = {New York, NY, USA},
volume = {60},
number = {6},
issn = {0001-0782},
url = {https://doi.org/10.1145/3065386},
doi = {10.1145/3065386},
journal = {Commun. ACM},
month = {may},
pages = {84–90},
numpages = {7}
}

@inproceedings{distributionshift,
author = {Ganin, Yaroslav and Lempitsky, Victor},
title = {Unsupervised domain adaptation by backpropagation},
year = {2015},
publisher = {JMLR.org},
booktitle = {Proceedings of the 32nd International Conference on International Conference on Machine Learning - Volume 37},
pages = {1180–1189},
numpages = {10},
location = {Lille, France},
series = {ICML'15}
}

@inproceedings{distillation,title	= {Distilling the Knowledge in a Neural Network},author	= {Geoffrey Hinton and Oriol Vinyals and Jeffrey Dean},year	= {2015},URL	= {http://arxiv.org/abs/1503.02531},booktitle	= {NIPS Deep Learning and Representation Learning Workshop}}

@inproceedings{textureshallow,
  title={Multi-feature co-learning for image inpainting},
  author={Lin, Jiayu and Wang, Yuan-Gen and Tang, Wenzhi and Li, Aifeng},
  booktitle={2022 26th International Conference on Pattern Recognition (ICPR)},
  pages={296--302},
  year={2022},
  organization={IEEE}
}

@article{threeways,
  author       = {Lukas Hoyer and
                  Dengxin Dai and
                  Yuhua Chen and
                  Adrian K{\"{o}}ring and
                  Suman Saha and
                  Luc Van Gool},
  title        = {Three Ways to Improve Semantic Segmentation with Self-Supervised Depth
                  Estimation},
  journal      = {CoRR},
  volume       = {abs/2012.10782},
  year         = {2020},
  url          = {https://arxiv.org/abs/2012.10782},
  eprinttype    = {arXiv},
  eprint       = {2012.10782},
  bibsource    = {dblp computer science bibliography, https://dblp.org}
}

@inproceedings{domainrandom1,
  title={Domain randomization and pyramid consistency: Simulation-to-real generalization without accessing target domain data},
  author={Yue, Xiangyu and Zhang, Yang and Zhao, Sicheng and Sangiovanni-Vincentelli, Alberto and Keutzer, Kurt and Gong, Boqing},
  booktitle={Proceedings of the IEEE/CVF international conference on computer vision},
  pages={2100--2110},
  year={2019}
}

@misc {frodobots_lab_2024,
	author       = { {FrodoBots Lab} },
	title        = { FrodoBots-2K (Revision 1abf1b8) },
	year         = 2024,
	url          = { https://huggingface.co/datasets/frodobots/FrodoBots-2K },
	doi          = { 10.57967/hf/3042 },
	publisher    = { Hugging Face }
}

\end{document}